\documentclass[a4paper]{article}

\usepackage[pages=all, color=black, position={current page.south}, placement=bottom, scale=1, opacity=1, vshift=5mm]{background}
\SetBgContents{
	\tt This work is shared under a \href{https://creativecommons.org/licenses/by-nc/4.0/deed.en}{CC BY-NC 4.0 license}.
}      

\usepackage[margin=1in]{geometry} 

\usepackage{amsmath}
\usepackage{amsthm}
\usepackage{amssymb}
\usepackage{booktabs}

\usepackage[ruled, lined, longend]{algorithm2e}

\usepackage[utf8]{inputenc}
\usepackage{hyperref}
\hypersetup{
	unicode,
	pdfauthor={Author One, Author Two, Author Three},
	pdftitle={A simple article template},
	pdfsubject={A simple article template},
	pdfkeywords={article, template, simple},
	pdfproducer={LaTeX},
	pdfcreator={pdflatex}
}

\usepackage{multirow}
\usepackage{caption}
\usepackage{subcaption}
\usepackage{amsmath}
\usepackage{amsfonts}
\usepackage{graphicx}

\theoremstyle{plain}
\newtheorem{theorem}{Theorem}

\theoremstyle{definition}
\newtheorem{definition}[theorem]{Definition}

\usepackage{graphicx, color}
\graphicspath{{fig/}}

\usepackage{lipsum}

\title{Uncovering Utility Functions from Observed Outcomes}
\author{Marta Grześkiewicz$^1$}

\date{
	$^1$University of Cambridge \\ \texttt{mg2192@cam.ac.uk}\\%
    [2ex]%
}

\begin{document}
	\maketitle
	
	\begin{abstract}
	   Determining consumer preferences and utility is a foundational challenge in economics. They are central in determining consumer behaviour through the utility-maximising consumer decision-making process. However, preferences and utilities are not observable and may not even be known to the individual making the choice; only the outcome is observed in the form of demand. Without the ability to observe the decision-making mechanism, demand estimation becomes a challenging task and current methods fall short due to lack of scalability or ability to identify causal effects. Estimating these effects is critical when considering changes in policy, such as pricing, the impact of taxes and subsidies, and the effect of a tariff. To address the shortcomings of existing methods, we combine revealed preference theory and inverse reinforcement learning to present a novel algorithm, Preference Extraction and Reward Learning (PEARL) which, to the best of our knowledge, is the only algorithm that can uncover a representation of the utility function that best rationalises observed consumer choice data given a specified functional form. We introduce a flexible utility function, the Input-Concave Neural Network which captures complex relationships across goods, including cross-price elasticities. Results show PEARL outperforms the benchmark on both noise-free and noisy synthetic data. 
	\end{abstract}
    \newpage
	\section{Introduction}
\label{section:introduction}

Preferences and utility are central concepts in economics and play a key role in determining consumer behaviour and market outcomes. Preferences refer to an ordering over possible choices based on an individual's needs and desires and under certain conditions, these preferences can be represented by a utility function. In micro-economic theory, the consumer is utility maximising. Utility is an intrinsic `reward' a consumer obtains when purchasing goods or services given their budget constraint. The outcome of this constrained optimisation problem is optimal consumption, or demand. Knowing the response of demand to changes in price and income is essential for estimating the effects of changes in policies which affect those variables. These responses can be used directly by firms to optimise their objectives, such as profit maximisation, sales maximisation, or indirectly by computing properties such as consumer surplus or elasticities. These properties are often used to determine economic policies such as the impact of taxes and subsidies, the effect of a tariff on prices, the valuation of non-market goods, and measuring the value of innovation \cite{BERRY20211, cohen2016uber}. 

Demand estimation from observational data is a difficult problem. The estimation of interest is the response of demand to changes in price or other factors, \textit{ceteris paribus}. The distinctive challenge appears when the presence of unobservable variables that affect both price and demand are acknowledged. The effects of such unobserved `demand shocks' arise from changes in tastes or preferences and lead to the problem of price endogeneity, where the effect of change in price on demand cannot be identified as it is confounded by the unobserved shocks. To measure demand or its elasticities, these demand shocks must be held fixed. This condition rules out common approaches for estimation with endogeneity, even in the hypothetical case of a single-good economy \cite{BERRY20211}. This is additionally complicated by the fact that demand for any one good will, in general, depend on the prices, observed characteristics and demand shocks of that good and all related goods. Allowing endogeneity of additional characteristics leads to more restrictions on econometric models.

Demand is often modeled with a discrete choice model \cite{McFa73, McFa1977}, where a consumer selects exactly one of the options available to them and the option can be one good, or more generally, a basket of goods. Discrete choice demand is often represented with a random utility model, where the consumer's utility vector corresponds to each available good and an `outside good' (choice of `none of the above') is drawn from a joint distribution, conditional on their characteristics and goods available. These models are frequently formulated using a parametric random utility specification \cite{McFa1977, hausman1978conditional, mcfadden1981econometric, berry1994estimating, berry1996airline, BERRY20211}. Such a formulation may allow the correlation of prices with the error, thus allowing for endogeneity but still requires specification of IVs, the requirements for which are heavy as an IV has to be specified for every product and every combination of products, which rapidly faces scalability issues. 

An alternative approach to uncovering preferences is using RP theory, a branch of micro-economics introduced by \cite{samuelson1974complementarity}. It uses observational data to obtain pairwise comparisons between chosen bundles of goods, which provides a ranking over all possible choices that can be represented by an ordinal utility function and is scale invariant. It also provides a method to verify that a dataset is consistent with the utility-maximising consumer with the condition that the ranking over choices inferred from the observations follows the axioms of revealed preference and there are no contradictions. For example, by the condition of transitivity, if $a$ is strictly preferred to $b$ and $b$ is strictly preferred to $c$, it cannot be that $c$ is strictly preferred to $a$. If the data is consistent, then a utility function which rationalises the observations can be inferred; with inconsistencies, the observed data can be adjusted by an efficiency index, such as Afriat's Index \cite{afriat1973}, to ensure it is consistent under the revealed preference axioms with adjustment. 

The proposed algorithm, Preference Extraction and Reward Learning (PEARL), is an iterative process based on revealed preferences and works in two steps: firstly, with the current estimate of the parameters of the specified utility function, we find the consumption which minimises expenditure while keeping utility at least as good as what is obtained under the observed consumption (the dual of the utility maximisation problem). This is computed by projected gradient optimisation which iteratively computes the optimal value given an initial starting point. Secondly, we compute the difference between the expenditure under the estimated optimal consumption and that of the observed consumption. This is the loss that is to be minimised by updating the parameters of the utility function which generated the estimated optimal consumption with gradient descent. Intuitively, the algorithm should return the observed consumption as the optimal consumption under the observed budget constraint, given the fitted utility function. 

The explicit treatment of prices as an input to the constraint of the utility maximisation problem means that, with changes in price, optimal consumption bundles that are consistent with the previous observations and the resulting utility function can be computed, thus isolating the effect of price on demand. Any deviations from this optimal consumption bundle can be viewed as the demand `shocks' and thus addressing the endogeneity problem found in demand estimation with regression. With subsequent observations, changing tastes and preferences can be shown by inconsistencies with the observed dataset under revealed preferences or by significant deviations from the estimated optimal output by the algorithm.

This approach is similar to the procedure in Inverse Reinforcement Learning (IRL) algorithms, where for the given estimate of the reward, the optimal action under a reward-maximising agent is found, and then the outcome of the optimal action with the observed outcomes is compared, often done by feature matching \cite{Ng2000}. However, unlike IRL methods, PEARL is centered around the constrained optimisation problem of utility maximisation under a budget constraint, and given that the data generating process is that of the utility-maximiser, revealed preference theory provides foundations for the guarantees of the existence of a utility function which rationalises the observed data. 

To address the requirement to specify a functional form of the utility function, we introduce multiple concave activation functions: \textit{concave-tanh}, \textit{concave-sigmoid} and \textit{concave-log}. The architecture of this neural network, the Input-Concave Neural Network (ICNN), is based on \cite{amos2017}. ICNN retains flexibility of the function while satisfying the properties of a utility function: it is monotonically increasing and concave with respect to the input. It allows own-price and cross-price elasticities to vary with prices and incomes, which can be estimated numerically. 

The performance of PEARL is shown on simulated data and real-world data. We investigate two settings: simulated data with noise-free observations and simulated data with noisy observations. In the first setting, we generate observations and show that PEARL can recover the true parameter values when the functional form of the utility function is known and we show that the ICNN is able to recover the utility function and predict demand and estimate elasticities with near-zero error. In the second setting, we perturb the generated observations with noise. We consider two types of noise: random noise on consumption and noise arising from endogeneity, both of which result in deviations from the optimal utility-maximising bundle. We compare the performance of PEARL to machine learning demand prediction methods, as presented in \cite{bajarietal2015}, and show that our methodology significantly out-performs the benchmark.

\section{Related Work} \label{section:related-work}

RP was first introduced by \cite{Samuelson1948ConsumptionTI} and further developed by \cite{houthakker1950revealed, afriat1967, diewert1973}; see \cite{varian1982nonparametric} and \cite{Tipoe2021} for an overview. Testing of RP axioms in observational data is discussed in \cite{SMEULDERS2019803}, and causes of violations in \cite{Tipoe2021}. 

Demand estimation is related to policy analysis, for which the standard econometric identification method is instrumental variables \cite{berry1994estimating, heckman1999local, heckman2005structural, angrist1995identification, mogstad2018identification, heckman2015causal}. Specification of instrumental variables is necessary as they are sources of exogenous variation which is required to overcome the effect of demand shocks on endogenous variables of price and demand. To be identified, at least $2k$ instrumental varialbes are required for $k$ goods, and often more are necessary to check for validity of instruments. PEARL overcomes these problems by implicitly holding the demand shocks across all goods fixed by keeping preferences constant under RP. When there are deviations from the utility-maximising optimal consumption, these are treated as demand shocks and de-coupled from the effect of price on demand.  

Other commonly used methods in the economics literature are using the matching assumption \cite{heckman1998characterizing}; the control function \cite{heckman1985alternative, telser1964iterative, blundell2003endogeneity}; and panel data approaches, such as separability and difference-in-difference. For demand prediction without explicit treatment of endogeneity, linear regression is still a popular benchmark model in demand forecasting \cite{ULRICH2021831}. Machine learning methods are shown as alternatives to the econometric model in \cite{CARBONNEAU20081140}, neural networks in \cite{ferreira2016study} and random forests have been applied in the context of load forecasting on the electricity market in \cite{lahouar2015}. Various methods for demand estimation for supermarket data are benchmarked in \cite{bajarietal2015}. All these methods fall short because they do not allow for identification for demand due to the lack of explicit treatment of endogeneity. See \cite{BERRY20211} for a detailed discussion. PEARL addresses these shortcomings by explicitly de-coupling the effect of changes in price on demand and demand shocks through solving the consumer's utility-maximisation problem where price is an input and demand is an output, thus addressing endogeneity. 

The structure of the proposed algorithm, PEARL, is similar to that found in the IRL literature; \cite{Ng2000} use a reward function linear in features and feature matching to evaluate the fitted reward function against the observations, while \cite{wulfmeier2015maximum} extend the method and fit a deep neural network as the reward function, while still evaluating with feature matching under the Maximum Entropy framework \cite{ziebart2008maximum}. These methods lack theoretical foundations of choice and decision-making and often have to solve for e.g. maximum entropy such that the reward function does not become degenerate and collapse to 0 everywhere. PEARL uses economic theory to underpin the mechanism of consumer choice, which informs the form of the utility function and avoids the problem of degeneration. 

Finally, \cite{Krishnamurthy2020} use Afriat's piece-wise linear utility function \cite{afriat1967} to solve the IRL problem in a probe and drone setting, where a drone/UAV or electromagnetic signal probes an `adversary' cognitive radar system which then estimates the kinematic coordinates of the drone/UAV using a Bayesian tracker and then adapts its mode (waveform, aperture, revisit time) dynamically using feedback control based on sensing the kinematic state of the drone. They apply RP to determine whether the radar is `rational' and use Afriat's piece-wise linear utility to find a utility function to predict the future behaviour of the cognitive radar. They use pairwise comparisons from RP under a utility-maximising and rational agent in a non-linear constraint setting (first presented by \cite{FORGES2009135}) to fit the piece-wise linear utility function given observations. In contrast, PEARL does not rely on constructing the ordered bundles and finding Afriat's numbers that result in an order-preserving piece-wise linear function. Instead, it introduces a model-agnostic algorithm that can fit any utility function given observations from a single time period, which either are or can be adjusted to be consistent under RP. This provides more accurate representations of the utility due to the input-concave nature of the function.

\section{Background} \label{section:background}

Consider an agent choosing between two bundles of $k$ goods denoted by the $k \times 1$ vectors $\textbf{x}_i$ and $\textbf{x}_j$, $\textbf{x}_i, \textbf{x}_j \in X \subseteq \mathbb{R}_{+}^k$, which denote the quantities that the agent chooses over the $k$ goods. If they choose $\textbf{x}_i$ over $\textbf{x}_j$, we say that $\textbf{x}_i$ is weakly preferred to $\textbf{x}_j$ and this is denoted as $\textbf{x}_i \succeq \textbf{x}_j$. In other words, $\textbf{x}_i$ is at least as good as $\textbf{x}_j$. Strict preference is denoted by $\textbf{x}_i \succ \textbf{x}_j$, while indifference is denoted by $\textbf{x}_i \sim \textbf{x}_j$. Following standard theory (see, for example,  \cite{varian2006intermediate}), the following three assumptions about preferences are made: firstly, they are complete, meaning that any two bundles can be compared; secondly, they are reflexive, meaning that any bundle is at least as good as itself ($\textbf{x}_i \succeq \textbf{x}_i$); and thirdly, they are transitive, meaning that if $\textbf{x}_i \succeq \textbf{x}_j$ and $\textbf{x}_j \succeq \textbf{x}_k$, then $\textbf{x}_i \succeq \textbf{x}_k$. While the first two assumptions are straightforward, transitivity is a strong assumption as it ensures that for any sets of options that are to be compared, there will be at least one best choice (the agent may be indifferent between multiple best choices). Further technical assumptions for the preferences to be `well-behaved' are made, such as monotonicity, convexity and local non-satiation. Note that convexity in preferences implies that averages (combinations of many goods) are preferred to extremes (consuming lots of one good).

Given such a preference ordering, by the representation theorem in \cite{debreu1954representation}, this can be described by a utility function indexed by $d$-dimensional parameter vector $\boldsymbol{\theta}$, denoted by $U_{\boldsymbol{\theta}}:\mathbb{R}^k_{+} \rightarrow \mathbb{R}_{+}$, where if $\textbf{x}_i \succeq \textbf{x}_j$, then $U_{\boldsymbol{\theta}}(\textbf{x}_i) \geq U_{\boldsymbol{\theta}}(\textbf{x}_j)$.\footnote{In the lottery space, for some probability attributed to the utility outcome, this is represented by the utility theorem in \cite{von1947theory}: $\mathbb{E}[U(\textbf{x}_i)] \geq \mathbb{E}[U(\textbf{x}_j)]$. Here, only the deterministic case is considered.} This utility function has to be monotonically increasing, (quasi-)concave. The agent is then faced with the following objective 
\begin{equation}
    \max_{\textbf{x}} U_{\boldsymbol{\theta}}(\textbf{x}) \textnormal{ s.t. } \textbf{p}^T_i \textbf{x} \leq m_i
\end{equation}
where the state is defined as $\textbf{s}_i = (\textbf{p}_i, m_i)$, where $\textbf{p}_i \in \mathbb{R}_{+}^k$ denotes the corresponding $k \times 1$ vector of prices. This corresponds to the `budget constraint', the total amount $m_i$ an agent can afford, given the current prices $\textbf{p}_i$. 

Suppose an agent is acting optimally under this objective and the observations $D = \{\textbf{x}_i, \textbf{p}_i, m_i\}_{i=1}^N$ are obtained. Neither the utility function nor its values are observed as they are intrinsic to the agent. The feasibility region, $O(\textbf{s}_i)$, is described as the alternative consumption choices that satisfy the condition $\textbf{x} \in O(\textbf{s}_i)$ $\in X$ if $\textbf{p}_i^T \textbf{x} \leq m_i$.  If an agent is observed to be acting optimally and always chooses what they most prefer, then it must be the case that $\textbf{x}_i \succeq \textbf{x}_j$, $\forall \textbf{x}_j \in O(\textbf{s}_i)$.

Under the linear constraint, for every observation, $\textbf{x}_i$ is \textit{strictly directly revealed preferred} to $\textbf{x}_j$, denoted $\textbf{x}_i P^D\textbf{x}_j$ if $\textbf{p}_i^T \textbf{x}_i > \textbf{p}_i^T\textbf{x}_j$. Then, the transitive closure is computed and $\textbf{x}_i$ is \textit{revealed preferred} to $\textbf{x}_j$, denoted $\textbf{x}_i R\textbf{x}_j$ if $\textbf{p}_i^T \textbf{x}_i \geq \textbf{p}_i^T \textbf{x}_k, \textbf{p}_k^T \textbf{x}_k \geq \textbf{p}_k^T \textbf{x}_l, \dots, \textbf{p}_m^T \textbf{x}_m \geq \textbf{p}_m^T \textbf{x}_j$ for some sequence of observations $(\textbf{x}_i, \textbf{x}_j, \dots, \textbf{x}_m)$. Given these revealed preference relations, a condition for consistency with revealed preferences called the Generalised Axiom of Revealed Preference (GARP) \cite{afriat1967} is defined:
\begin{definition}[GARP] 
    If $\textbf{x}_i R \textbf{x}_j$, then $\textbf{x}_j P^D \textbf{x}_i$ cannot be true and $\textbf{x}_i R \textbf{x}_j \text{ implies } \textbf{p}_j^T \textbf{x}_i \leq \textbf{p}_j^T \textbf{x}_j$.
\end{definition}

Moreover, under Afriat's Theorem \cite{afriat1967}, if the data is consistent under GARP, then there is an increasing, monotonic, continuous and concave utility function that rationalises this data. GARP-consistency is a testable property of the data set and we can check whether a data set is consistent under GARP with algorithms of transitive closure, such as that presented in \cite{Warshall1962ATO} which has computational complexity $O(N^3)$, combined with an algorithm that checks for cycles in a directed graph, such as Depth-First-Search \cite{cormen2022introduction}. \cite{Krishnamurthy2020} introduce a statistical test to check whether a data set is GARP-consistent even in the presence of noise from measurement error. \cite{Tipoe2021} review various reasons for the violation of GARP and their remedies. 

In the case when the data set is not GARP-consistent, `goodness-of-fit' indices can be obtained, for example Afriat's index \cite{afriat1973}, Varian's index \cite{VARIAN1990125} and the Houtman-Maks index \cite{HoutmanMaks1985}; the computational complexity of these indicies is discussed in \cite{SMEULDERS2019803}. Of these, Afriat’s index is the simplest, where a value $0 \leq \varepsilon \leq 1$ is introduced for all observations. It relaxes the RP relations with the following adjustment: 
\begin{equation}
\begin{split}
    &\text{ if } \varepsilon \textbf{p}_i^T \textbf{x}_i \geq \textbf{p}_i^T \textbf{x}_j, \text{ then } \textbf{x}_i R^D(\varepsilon) \textbf{x}_j \\
    &\text{ if } \varepsilon \textbf{p}_i^T \textbf{x}_i > \textbf{p}_i^T \textbf{x}_j, \text{ then } \textbf{x}_i P^D(\varepsilon) \textbf{x}_j \\
\end{split}
\end{equation}
where $R(\varepsilon)$ and $P(\varepsilon)$ represent the transitive closures of $R^D(\varepsilon)$ and $P^D(\varepsilon)$, respectively.

\begin{definition}
    For a given $\varepsilon$, with $0 \leq \varepsilon \leq 1$, a dataset $D = \{\textbf{p}_i, \textbf{x}_i\}_{i=1}^N$ satisfies GARP($\varepsilon$) if and only if, for each pair of distinct bundles, $\textbf{x}_i, \textbf{x}_j, i, j =1,\dots, N; i\neq j$, if $\textbf{x}_i R(\varepsilon) \textbf{x}_j$, then it cannot be that $\textbf{x}_j P^D(\varepsilon)\textbf{x}_i$.
\end{definition}

\section{Methodology} \label{section:methodology}

In this section, we present the PEARL algorithm. Under RP theory, the observed data has to be GARP-consistent for a concave, increasing, monotonic utility function to exists. As such, we first check for this consistency, and if it is not consistent, we adjust the data. We then present the algorithm for recovering the utility function, given a specified functional form. Finally, we present a numerical method for maximising the recovered utility given state variables.

\subsection{GARP-consistency} \label{section:PEARL-garp}

We verify that the set of observations $D$ is consistent with optimality conditions under RP theory. The $N \times N$ binary relation matrix $\textbf{R}$ is constructed with elements $i, j = 1, \dots, N$, 
\begin{equation}
    \textbf{R}[i,j] = \begin{cases}
         1, & \text{if } \textbf{p}_i^T \textbf{x}_i \geq \textbf{p}_i^T \textbf{x}_j  \text{, and } i \neq j \\
         0, & \text{otherwise. }
    \end{cases}
\end{equation}
The matrix $\textbf{R}$ defines a directed graph such that every node is an observation and every edge represents whether the observation is preferred to the other. GARP inconsistencies are found by constructing the transitive closure graph \cite{Warshall1962ATO} which is then checked for cycles using e.g. the Depth-First-Search algorithm; see, for example, \cite{cormen2022introduction}. If inconsistencies are present, then the data is adjusted by Afriat's efficiency index, $0<\varepsilon\leq1$, which is defined as the maximum value of $\varepsilon$ such that the adjusted dataset $D' = \{\varepsilon \textbf{x}_i, \textbf{p}_i, m_i  \}_{i=1}^N $ has no inconsistencies. This ensures the binary relation under transitive closure is acyclic. The value of $\varepsilon$ is approximated by binary search. 

\subsection{Recovering Utility} \label{section:PEARL-grad}

With GARP-consistent observations and under Afriat's theorem, there exists an increasing, monotonic, concave utility function that rationalises the data. We introduce our algorithm, Preference Extraction and Reward Learning (PEARL), which recovers a representation of the underlying utility function given a set of observations and a specified functional form which satisfies the properties of a utility function.

Since we observe only the outcome of consumer choice in terms of consumption and expenditure, $\textbf{x}_i$ and $m_i$, respectively, PEARL compares the observed values $m_i$ to computed optimal $m$-values, $\hat{m}(\textbf{p}_i, u_i)$ by a loss function. The $m$-values are solutions to the dual problem of the utility maximisation objective, also known as the money metric utility \cite{samuelson1974complementarity}. It is defined as 
\begin{equation} \label{eq:money-metric}
    \hat{m}(\textbf{p}_i, u_i) = \min_{\textbf{x}} \textbf{p}_i^T \textbf{x} \textnormal{ s.t. } U_{\theta}(\textbf{x}) \geq u_i 
\end{equation}
where $\hat{m}(\textbf{p}_i, u_i)$ is the minimum cost payable in order to achieve the utility level $u_i$ for some parameterised utility function $U_{\boldsymbol{\theta}}$. Here and in the following sections, the utility function is indexed by the unknown parameter vector $\boldsymbol{\theta}$, which is object of interest in recovering the utility function. These values are compared by the L1 loss, given by  
\begin{equation} \label{eq:loss}
	L(\boldsymbol{\theta}) = \sum_{i=1}^N L_i(\boldsymbol{\theta}) \text{ where } L_i(\boldsymbol{\theta}) = |(\hat{m}(\textbf{p}_i, u_i) - m_i)|
\end{equation}
PEARL estimates the parameters $\boldsymbol{\theta}$ in two iterative stages. First, for every observation $i=1, \dots, N$ we find the action $\hat{\textbf{h}}_i$ which, under the current utility function $U_{\hat{\theta}_t}$, is at least as good as the observed $\textbf{x}_i$ and which minimises expenditure, given by 
\begin{equation} \label{eq:money-metric-objective}
    \hat{\textbf{h}}_i = \arg \min_{\textbf{x}} \textbf{p}_i^T \textbf{x} \text{ s.t. } U_{\boldsymbol{\hat{\theta}}}(\textbf{x}) \geq u_i.
\end{equation} 
where the result $\hat{\textbf{h}}_i$ corresponds to the optimal (Hicksian) demand (see, for example, \cite{varian2006intermediate}) where the utility level is at least $u_i$. The second stage uses these values to compute the loss and its gradients, and updates the parameters with gradient descent. 

We now present the algorithm in more detail. The first stage computes the corresponding $\hat{\textbf{h}}_i$ for every observation $i=1,\dots,N$ using a numerical method with projections. The method of projected gradients is used as with the constraint set of equation \ref{eq:money-metric-objective} defined to be $C_i = \{\textbf{x} : U_{\boldsymbol{\hat{\theta}}}(\textbf{x}) \geq u_i \}$, it is known to be a convex set because $U_{\boldsymbol{\hat{\theta}}}$ is by definition concave in $\textbf{x}$. The objective to find the money metric utility can be written, for $i = 1, \dots, N$, as $\hat{\textbf{h}}_i = \arg \min_{\textbf{x} \in C_i} \textbf{p}_i^T\textbf{x}$.
The projected gradient calculations are then, with learning rate $\alpha$, firstly to perform a gradient descent step 
\begin{equation}
    \hat{\textbf{h}}_{i, t+\frac{1}{2}} = \hat{\textbf{h}}_{i, t} + \alpha  \frac{\textbf{p}_i}{|\textbf{p}_i|} 
\end{equation}
where $\hat{\textbf{h}}_{i,t+\frac{1}{2}}$ denotes the intermediate step between $\hat{\textbf{h}}_{i,t}$ and $\hat{\textbf{h}}_{i,t+1}$. Then we compute the projection back onto the constraint set by finding the minimum distance from the intermediate point onto the constraint set
\begin{equation}
    \hat{\textbf{h}}_{i, t+1} = \arg \min_{\textbf{y} \in C_i} || \textbf{y} -\hat{\textbf{h}}_{i, t+\frac{1}{2}} ||
\end{equation}
Setting up equation \ref{eq:money-metric-objective} as a Lagrangian, by the Karush-Kuhn-Tucker (KKT) conditions \cite{karush1939minima, kuhn1951nonlinear}, the constraint will bind such that $U_{\boldsymbol{\hat{\theta}}}(\hat{\textbf{x}}) = u_i$ and 
\begin{equation}
    \frac{\textbf{p}_i}{|\textbf{p}_i|} = \frac{\partial U_{\boldsymbol{\hat{\theta}}}(\hat{\textbf{h}}_{i, t+\frac{1}{2}})/\partial \hat{\textbf{h}}_{i, t+\frac{1}{2}}}{|\partial U_{\boldsymbol{\hat{\theta}}}(\hat{\textbf{h}}_{i, t+\frac{1}{2}})/\partial \hat{\textbf{h}}_{i, t+\frac{1}{2}}|}
\end{equation}
As such, we take $\textbf{y}$ as the point after a step is taken in the direction of normalised gradient of the utility function with respect to $\hat{\textbf{h}}_{i,t+\frac{1}{2}}$ by the amount $U_{\boldsymbol{\hat{\theta}}}(\hat{\textbf{x}}) - u_i$, as an approximation of the shortest distance for small enough $\alpha$. Hence, 
\begin{equation}
    \textbf{y} = \hat{\textbf{h}}_{i, t+\frac{1}{2}} + (u_i - U_{\boldsymbol{\hat{\theta}}}(\hat{\textbf{h}}_{i, t+\frac{1}{2}})) \frac{\partial U_{\boldsymbol{\hat{\theta}}}(\hat{\textbf{h}}_{i, t+\frac{1}{2}})/\partial \hat{\textbf{h}}_{i, t+\frac{1}{2}}}{|\partial U_{\boldsymbol{\hat{\theta}}}(\hat{\textbf{h}}_{i, t+\frac{1}{2}})/\partial \hat{\textbf{h}}_{i, t+\frac{1}{2}}|} 
\end{equation}
and so
\begin{equation}
    \hat{\textbf{h}}_{i, t+1} = \hat{\textbf{h}}_{i, t} + \alpha   \frac{\textbf{p}_i}{|\textbf{p}_i|} + (u_i - U_{\boldsymbol{\hat{\theta}}}( \hat{\textbf{h}}_{i, t+\frac{1}{2}}))\frac{\partial U_{\boldsymbol{\hat{\theta}}}(\hat{\textbf{h}}_{i, t+\frac{1}{2}})/\partial \hat{\textbf{h}}_{i, t+\frac{1}{2}}}{|\partial U_{\boldsymbol{\hat{\theta}}}(\hat{\textbf{h}}_{i, t+\frac{1}{2}})/\partial \hat{\textbf{h}}_{i, t+\frac{1}{2}}|}
\end{equation}
This is summarised in Algorithm \ref{algo:proj-grad}.

\begin{algorithm}[H]
\caption{Computing Money Metric Utility}\label{algo:proj-grad}
\KwIn{$\textbf{p}_i, u_i, i = 1, \dots, N , U_{\hat{\boldsymbol{\theta}}}$, learning rate $\alpha$, initial values $\hat{\textbf{x}}_{i, 0}$, number of iterations $T_m$.}
\KwOut{$\hat{\textbf{h}}_i$, $\hat{m}_i$, $i=1, \dots, N$.}
\For{$t = 0, \dots, T_m-1$}{
Compute intermediate step 
    \begin{equation}
        \hat{\textbf{h}}_{i, t+\frac{1}{2}} = \hat{\textbf{h}}_{i, t} + \alpha \frac{\textbf{p}_i}{|\textbf{p}_i|}
    \end{equation} \\
Compute the projected gradient and update 
\begin{equation*}
        \hat{\textbf{h}}_{i, t+1} = \hat{\textbf{h}}_{i, t} + \alpha   \frac{\textbf{p}_i}{|\textbf{p}_i|} + (u_i - U_{\boldsymbol{\hat{\theta}}}( \hat{\textbf{h}}_{i, t+\frac{1}{2}}))\frac{\partial U_{\boldsymbol{\hat{\theta}}}(\hat{\textbf{h}}_{i, t+\frac{1}{2}})/\partial \hat{\textbf{h}}_{i, t+\frac{1}{2}}}{|\partial U_{\boldsymbol{\hat{\theta}}}(\hat{\textbf{h}}_{i, t+\frac{1}{2}})/\partial \hat{\textbf{h}}_{i, t+\frac{1}{2}}|}
\end{equation*} \\
}
Set $\hat{\textbf{h}}_i = \hat{\textbf{h}}_{i, T_m}$; $\hat{m}_i = \textbf{p}_i^T\hat{\textbf{h}}_i$.
\end{algorithm}

Given these estimates of $\hat{m}_i$ and $\hat{\textbf{h}}_i$,  the loss of how far away the cardinal value of the $m$-value is from the estimated and the observed is computed, given the observed prices $\textbf{p}_i$. The parameters of the utility function are updated by gradient descent, with the gradient computation given below. Notice that with an assumed strictly concave utility function, the values $\hat{m}_i$ and $\hat{\textbf{h}}_i$ are uniquely determined. The L1 loss is given by equation \ref{eq:loss} and the parameters of the utility function, $\boldsymbol{\theta}$, are optimised by gradient descent. By the chain rule, 
\begin{equation}
    \frac{\partial L_i(\hat{\boldsymbol{\theta}})}{\partial\boldsymbol{\theta}} = \left( \frac{\partial  L_i(\boldsymbol{\theta})}{\partial \hat{m}_i} \right) \left( \frac{\partial \hat{m}_i}{\partial \hat{\textbf{h}}_i} \right)^T \left( \frac{\partial \hat{\textbf{h}}_i}{\partial \boldsymbol{\theta}}\right)
\end{equation}
where $\frac{\partial \hat{\textbf{h}}_i}{\partial \hat{\boldsymbol{\theta}}}$ denotes the $k \times d$ matrix with elements as derivatives of $\hat{\textbf{h}}_i$ with respect to the elements of $\boldsymbol{\theta}$. With the abuse of notation, the derivative of the loss function given in equation \ref{eq:loss} is given by
\begin{equation}
    \frac{\partial L_{i}(\boldsymbol{\theta})}{\partial \hat{m}_i} = \frac{|(\hat{m}_i - m_i)|}{(\hat{m}_i - m_i)} 
\end{equation}
for $\hat{m}_i \neq m_i$, and we set the value at $0$ otherwise. Additionally, from equation \ref{eq:money-metric},
\begin{equation}
	\left(\frac{\partial m_i}{\partial \hat{\textbf{h}}_i}\right)^T \left( \frac{\partial \hat{\textbf{h}}_i}{\partial \boldsymbol{\theta}} \right) = \textbf{p}_i \frac{\partial \hat{\textbf{h}}_i}{\partial \boldsymbol{\theta}}.
\end{equation}
To find an expression for $\frac{\partial \hat{\textbf{h}}_i}{\partial \hat{\boldsymbol{\theta}}}$, the Lagrangian is constructed for the money-metric utility for the PEARL algorithm, denoted as $\mathcal{L}$, as 
\begin{equation}
    \mathcal{L}(\hat{\textbf{h}}_i, \hat{\boldsymbol{\theta}}; \textbf{p}_i, \textbf{x}_i) = \textbf{p}_i^T \hat{\textbf{h}}_i - \lambda (U_{\hat{\boldsymbol{\theta}}}(\hat{\textbf{h}}_i) - U_{\hat{\boldsymbol{\theta}}}(\textbf{x}_i)) 
\end{equation}
where $\lambda$ is the Lagrange multiplier. The first order conditions are given by 
\begin{equation} \label{eq:lagrange}
	\frac{\partial \mathcal{L}}{\partial \hat{\boldsymbol{\theta}}} = \textbf{p}_i \frac{\partial \hat{\textbf{h}}_i}{\partial \hat{\boldsymbol{\theta}}} - \lambda \left( \frac{\partial U_{\hat{\boldsymbol{\theta}}}(\hat{\textbf{h}}_i)}{\partial \hat{\boldsymbol{\theta}}} - \frac{\partial U_{\hat{\boldsymbol{\theta}}}(\textbf{x}_i)}{\partial \hat{\boldsymbol{\theta}}}\right) - \frac{\partial\lambda}{\partial\boldsymbol{\hat{\boldsymbol{\theta}}}} \left( U_{\hat{\boldsymbol{\theta}}}(\hat{\textbf{h}}_i) - U_{\hat{\boldsymbol{\theta}}}(\textbf{x}_i) \right)
\end{equation}
\begin{equation} \label{eq:rp_icb_lagrange_x}
	\frac{\partial \mathcal{L}}{\partial \hat{h}_{i,j}} = p_{i,j} - \lambda \frac{\partial U_{\hat{\boldsymbol{\theta}}}(\hat{\textbf{h}}_i)}{\partial \hat{h}_{i,j}} 
\end{equation}
where $j$ refers to the $j^{th}$ element of $\hat{\textbf{h}}_i$ and $\textbf{p}_i$, $j=1, \dots, k$. At optimum, from Equation \ref{eq:rp_icb_lagrange_x}, the set of expressions is obtained for $j=1, \dots, k$, 
\begin{equation}
    \hat{\lambda} = \frac{p_{i,j}}{\partial U_{\hat{\boldsymbol{\theta}}} (\hat{\textbf{h}}_i) /\partial \hat{h}_{i,j}}
\end{equation}
which should be equal at optimum for every $j$. As an approximation during training it is set as
\begin{equation}
    \hat{\lambda} = k^{-1}  \textbf{p}_i^T \cdot \left( \frac{\partial U_{\hat{\boldsymbol{\theta}}}}{\partial \hat{\textbf{h}}_i}\right)^{-1}
\end{equation}
Also from Equation \ref{eq:lagrange}, at optimum, since $\hat{\textbf{x}}_i$ is the argument which maximises the utility maximisation objective for $i=1, \dots, N$ observations, $(U_{\hat{\boldsymbol{\theta}}}(\hat{\textbf{x}}_i) - U_{\hat{\boldsymbol{\theta}}}(\textbf{x}_i))=0$ is satisfied. So, 
\begin{equation}
	\textbf{p} \frac{\partial \hat{\textbf{h}}_i}{\partial \hat{\boldsymbol{\theta}}} = \hat{\lambda} \left( \frac{\partial U_{\hat{\boldsymbol{\theta}}}(\hat{\textbf{h}}_i)}{\partial \hat{\boldsymbol{\theta}}} - \frac{\partial U_{\hat{\boldsymbol{\theta}}}(\textbf{x})}{\partial \hat{\boldsymbol{\theta}}}\right). 
\end{equation}
Therefore, for $\hat{m}_i \neq m_i$ and $0$ otherwise,
\begin{equation} \label{eq:rp_icb_gradient}
    \frac{\partial L_i}{\partial \hat{\boldsymbol{\theta}}} = \frac{|(\hat{m}_i - m_i)|}{(\hat{m}_i - m_i)}  k^{-1}  \textbf{p}_i^T \cdot \left( \frac{\partial U_{\hat{\boldsymbol{\theta}}}}{\partial \hat{\textbf{h}}_i}\right)^{-1} \left( \frac{\partial U_{\hat{\boldsymbol{\theta}}}(\hat{\textbf{h}}_i)}{\partial \hat{\boldsymbol{\theta}}} - \frac{\partial U_{\hat{\boldsymbol{\theta}}}(\textbf{x}_i)}{\partial \hat{\boldsymbol{\theta}}}\right)
\end{equation}
The methodology is summarised by Algorithm \ref{algo:rp-bandit}.
\begin{algorithm}[h]
\caption{PEARL}\label{algo:rp-bandit}
\KwIn{$D = \{ (\textbf{x}_i, \textbf{p}_i, m_i ) \}_{i=1}^N$, initial parameters $\hat{\boldsymbol{\theta}}_0$, number of iterations $T$}
\KwOut{Utility function $U_{\hat{\boldsymbol{\theta}}_t}(\cdot)$}
Construct revealed preference graph by binary preference relation and transitive closure and check for GARP-consistency (acyclicality)\;
\For{$t = 1, \dots, T$}{
Set $u_i = U_{\hat{\boldsymbol{\theta}}_t}(\textbf{x}_i)$, for $i = 1, \dots, N$\;
Compute $\hat{\textbf{h}}_i$, for $i = 1, \dots, N$ by Algorithm \ref{algo:proj-grad} and set
    $\hat{m}_i = \textbf{p}^T_i \hat{\textbf{h}}_i$\;
Compute loss $L(\hat{\boldsymbol{\theta}}_t) = \sum_{i=1}^N L_i(\hat{\boldsymbol{\theta}}_t)$. \;
Update parameters $\hat{\boldsymbol{\theta}}_t$ by gradient descent, where for $\hat{m}_i \neq m_i$ and $0$ otherwise,
\begin{equation*}
        \frac{\partial L_i}{\partial \hat{\boldsymbol{\theta}}_t} = \frac{|(\hat{m}_i - m_i)|}{(\hat{m}_i - m_i)}  k^{-1}  \textbf{p}_i^T \cdot \left( \frac{\partial U_{\hat{\boldsymbol{\theta}}_t}}{\partial \hat{\textbf{h}}_i}\right)^{-1} \left( \frac{\partial U_{\hat{\boldsymbol{\theta}}_t}(\hat{\textbf{h}}_i)}{\partial \hat{\boldsymbol{\theta}}_t} - \frac{\partial U_{\hat{\boldsymbol{\theta}}_t}(\textbf{x}_i)}{\partial \hat{\boldsymbol{\theta}}_t}\right)
\end{equation*}
    }
Set $\hat{\boldsymbol{\theta}} = \hat{\boldsymbol{\theta}}_T$\;
\end{algorithm}

\subsection{Maximising Utility} \label{section:maximising-utility}

Given a representation of the utility function, $U_{\hat{\boldsymbol{\theta}}}$, obtained by PEARL, the counterfactual prediction for $\hat{\textbf{x}}$ under some given environmental variables $(\textbf{p}, m)$ is obtained by solving the consumer's utility-maximising objective ($\hat{\textbf{x}}^M = \arg \max_{\textbf{x}} U_{\hat{\boldsymbol{\theta}}}(\textbf{x}) \text{ s.t. } \textbf{p}^T\textbf{x} \leq m$), which provides an estimate of (Marshallian) demand (see, for example, \cite{varian2006intermediate}). We present a method to compute this numerically by iteration. An initial point is chosen and projected onto the feasibility set, which, since the constraint is linear, is a simple scaling such that the constraint holds with equality $\hat{\textbf{z}}_{i, t} = \hat{\textbf{x}}_i^M \frac{m_i}{\textbf{p}_i \cdot \hat{\textbf{x}}_i^M}$. The point is then moved along the constraint by the update rule 
\begin{equation}
    \hat{\textbf{z}}_{i, t+1} =  \hat{\textbf{z}}_{i, t} + \frac{\partial U_{\hat{\boldsymbol{\theta}}}(\hat{\textbf{z}}_{i, t}) / \partial \hat{\textbf{z}}_{i, t}}{|\partial U_{\hat{\boldsymbol{\theta}}}(\hat{\textbf{z}}_{i, t}) / \partial \hat{\textbf{z}}_{i, t}|} - |\textbf{p}|^T \cdot |\partial U_{\hat{\boldsymbol{\theta}}}(\hat{\textbf{z}}_{i, t}) / \partial \hat{\textbf{z}}_{i, t}| |\textbf{p}| 
\end{equation}
Notice that at optimum, under the KKT conditions, $\textbf{p}$ and $\frac{\partial U_{\hat{\boldsymbol{\theta}}}(\hat{\textbf{z}}_{i, t}) }{ \partial \hat{\textbf{z}}_{i, t}}$ will be equivalent. The procedure is summarised in Algorithm \ref{algo:max-u}. 

\begin{algorithm}[ht]
\caption{Maximise Utility}\label{algo:max-u}
\KwIn{$U_{\boldsymbol{\hat{\theta}}}, (\textbf{p}_i, m_i)_{i=1}^N$, number of iterations $T$}
\KwOut{$\{\hat{\textbf{x}}\}_{i=1}^N, \{\hat{m_i}\}_{i=1}^N$}
Initialise $\hat{\textbf{x}}_i^M$, for $i = 1, \dots, N$\;
Project $\hat{\textbf{x}}_i^M$ onto the constraint set by setting $\textbf{z}_i = \hat{\textbf{x}}_i^M \frac{m_i}{\textbf{p}_i \cdot \hat{\textbf{x}}_i^M}$ \\

\For{$t = 1, \dots, T$}{

$\hat{\textbf{z}}_{i, t+\frac{1}{2}} = \hat{\textbf{z}}_{i, t} + \frac{\partial U_{\boldsymbol{\hat{\theta}}}(\hat{\textbf{z}}_{i, t}) / \partial \hat{\textbf{z}}_{i, t}}{|\partial U_{\boldsymbol{\hat{\theta}}}(\hat{\textbf{z}}_{i, t}) / \partial \hat{\textbf{z}}_{i, t}|} - |\textbf{p}_i|^T \cdot |\partial U_{\boldsymbol{\hat{\theta}}}(\hat{\textbf{z}}_{i, t}) / \partial \hat{\textbf{z}}_{i, t}| |\textbf{p}_i| $ \\

Project $\hat{\textbf{z}}_{i, t+\frac{1}{2}}$ onto the constraint set by setting $\hat{\textbf{z}}_{i, t+1} = \hat{\textbf{z}}_{i, t+\frac{1}{2}} \frac{m_i}{\textbf{p}_i \cdot \hat{\textbf{z}}_{i, t+\frac{1}{2}}}$ \\
}

Set $\hat{\textbf{x}}_i^M = \hat{\textbf{z}}_{i,T}$
\end{algorithm}

\subsubsection{Adjusting for GARP-inconsistencies} \label{section:garp-consistency}
When the dataset is adjusted with $\varepsilon$ to achieve GARP-consistency, the inverse of the adjustment is applied to the outcomes of the utility maximisation problem. To achieve this, the epsilon is included in the utility condition when minimising the expenditure, $u_i = U_{\hat{\boldsymbol{\theta}}}(\varepsilon \textbf{x}_i)$ and in the transformation of the optimal value under the current utility function
\begin{equation}
    \hat{\textbf{x}}_i = \frac{1}{\varepsilon } \arg \min_{\textbf{x}} \textbf{p}^T_i \textbf{x} \text{ s.t. } U_{\hat{\boldsymbol{\theta}}}(\textbf{x}) \geq u_i
\end{equation}

\subsubsection{Numerical Estimation of Price Elasticities} \label{section:elasticities}

Elasticities are defined as the relative change in demand due to a change in price (see, for example \cite{varian2006intermediate}). The expression in equation \ref{eq:elasticities}, based on \cite{kocoska2009numerical}, gives a numerical approximation for (uncompensated) price elasticity of demand for good $i$ with respect to the change in price of good $j$, when the partial derivative cannot be computed directly. 
\begin{equation} \label{eq:elasticities}
    e_{i,j} = \frac{\partial x^i}{\partial p^j} \approx \frac{p^j}{x^i} \frac{\Delta x^i}{\Delta p^j}
\end{equation}
where when $i = j$ we compute own-price elasticity, and $i \neq j$ relating to cross-price for $i, j = 1, \dots, k$. The price vector is perturbed around the original vector $\textbf{p}$ in both plus and minus, per element $j$. n
Notice that these elasticities are not partitioned; the Hicks-Slutsky partition \cite{dixon1980} can also be computed with the corresponding constrained optimisation problems to compute both compensated and uncompensated price elasticity of demand. 

\subsection{Utility Functions} \label{subsection:utility-functions}

To illustrate the performance of the method, two functional forms of utility functions are considered: Cobb-Douglas (CD), and Input-Concave Neural Network (ICNN). The first is used in both simulating data and fitting, whilst the last, the ICNN is considered for only fitting the utility function using PEARL and for computing elasticties. 

\subsubsection{Cobb-Douglas}
The Cobb-Douglas utility function \cite{cobbdouglas} is a widely used mathematical representation to describe consumer utility with constant elasticities of substitution. The $k$-variable case is described as 
\begin{equation} \label{eq:cobb-douglas}
    U^{CD}_{\boldsymbol{\theta}}(\textbf{\textbf{x}}) = \prod_j^k x_j^{\theta_j} \text{ s.t. } \sum_j^k \theta_j = 1
\end{equation}
where $x_j$ refers to the $j$-th good (element) of a bundle of goods $\textbf{x}$, $j = 1, \dots, k$. It is one of the simplest representations of utility and we use this as the basic case to show the effectiveness of PEARL at recovering utility functions.

\subsubsection{Input-Concave Neural Network} \label{section:icnn}

The performance of the algorithm is investigated by assuming that the utility function is a complex parametric function, expressed by a neural network. As per the requirements of the algorithm, the function has to be input-concave and increasing, i.e. the derivative with respect to the input is positive and monotonically decreasing, and the second derivative with respect to the input is negative. 

The implementation of the Input-Concave Neural Network (ICNN) is based on the Input-Convex Neural Network, presented in \cite{amos2017}. A neural network is defined over inputs $\{\textbf{x}\}_{i=1}^N$ for $\ell=0,\dots,L$ fully connected (dense) layers. The output of each layer is
\begin{equation}
    z_{\ell+1}(\textbf{x}_i)  = h_{\ell} \left( W^{(z)}_{\ell} z_{\ell} + W^{(\textbf{x})}_{\ell} \textbf{x}_i + b_{\ell} \right); U^{ICNN}_{\theta}(\textbf{x}_i) = z_L(\textbf{x}_i) 
\end{equation}
where, with $z_0$, $W^{(z)}_0=0$,  $z_{\ell}$ denotes the layer activation functions, $h_{\ell}$ are the non-linear activation functions and $\boldsymbol{\theta} = \{ W_{0:\ell-1}^{(\textbf{x})}, W_{1:\ell-1}^{(z)}, b_{0:\ell-1} \}$ are the parameters. The function $U^{ICNN}_{\boldsymbol{\theta}}(\textbf{x}_i)$ is concave in $\textbf{x}_i$ provided that all $W^{(z)}_{1:L-1}$ are non-negative and all functions $h$ are concave and non-decreasing, following the convex proposition of the same nature presented in \cite{amos2017}. The proof follows from the fact that non-negative sums of concave functions are also concave and that the composition of two concave functions are also concave. The layer activation function weights are enforced to be non-negative by constructing an activation function to be concave.
Multiple candidates for concave activation functions are considered: 
\paragraph{Concave-tanh} Adapting the hyperbolic tangent, we obtain the `concave-tanh' activation function, given by
\begin{equation}
h(\textbf{x}) = 
    \begin{cases}
        \tanh(\textbf{x}), &\textbf{x} \geq \textbf{0} \\
        \textbf{x}, &\textbf{x} < \textbf{0}. 
    \end{cases}
\end{equation}

\paragraph{Concave-sigmoid} Adapting the sigmoid function, we obtain the `concave-sigmoid' activation function, given by
\begin{equation}
h(\textbf{x}) = 
    \begin{cases}
        \frac{1}{1+e^{-\textbf{x}}}, &\textbf{x} \geq \textbf{0} \\
        \frac{1}{4}\textbf{x}+\frac{1}{2}, &\textbf{x} < \textbf{0}. 
    \end{cases}
\end{equation}

\paragraph{Concave-log} Adapting the natural log, we obtain the `concave-log' activation function, where $\delta > 0$ is an offset necessary to avoid the asymptote at $\textbf{x}=0$, given by 
\begin{equation}
h(\textbf{x}) = 
    \begin{cases}
        \ln(\textbf{x} + \delta), &\textbf{x} > \textbf{0} \\
        \frac{1}{\delta}\textbf{x}+\ln{\delta}, &\textbf{x} \leq \textbf{0}. 
    \end{cases}
\end{equation}

\subsection{Pre-training on the Afriat numbers}

We can pre-train the network as a supervised learning problem given utility values obtained from Afriat's piecewise-linear utility function. Theorem says that if the data is GARP-consistent, then we can get a piece-wise linear function, such that there exist numbers $U_i, \lambda_i > 0$ such that $U_i \leq U_j + \lambda_j p_j(x_i - x_j)$ for $i, j = 1, \dots, n$. These numbers are found in the iterative procedures detailed in \cite{varian1982nonparametric}. 

\subsection{Implementation Details}

The training and model parameters have been optimised across multiple options and the resulting values are presented. The ICNN model is trained using the Adam optimiser \cite{kingma2017adammethodstochasticoptimization} with an epsilon value of $1.0e-5$ and weight decay of $5.0e-4$. The initial learning rate is set at $1.0e-3$, decreasing over time with a Exponential Decay scheduler with a final learning rate of $1.0e-8$ and a decay rate of . We set the batch size at $128$. We train for $1,000$ steps for Algorithm \ref{algo:proj-grad} at every iteration with a learning rate of $1.0e-3$. When the utility function is set to CD, it is trained for $1,000$ epochs. Unless otherwise specified, when the utility function is set as ICNN, we construct the network with $3$ layers and $k$ hidden units in each layer, where $k$ is the number of dimensions of the input $\textbf{x}$. This ensures that there is no dimensionality reduction whilst the number of parameters in the network is kept as small as possible. The architecture of the network is as described in section \ref{section:icnn}, with the addition of batch normalisation before the first dense layer and dropout after the first and before the last dense layers, neither of which affect the input-concavity of the network. Unless otherwise stated, the activation function is taken to be \textit{concave-log} with $\delta =0.01$. The model is trained for $10,000$ epochs due to the larger number of parameters.

\section{Simulations} \label{section:simulations}

In this section, we illustrate the performance of PEARL on simulated data in two cases: firstly, with Cobb-Douglas as ground-truth utility with no noise; secondly, with Cobb-Douglas as ground-truth utility but the observations are affected by noise. We compare prediction error to machine learning regression methods benchmarked in \cite{bajarietal2015}.

\subsection{Simulations without Noise}

To asses the performance of PEARL on observations without noise, the observations are simulated as consumption choices made by a consumer maximising their utility function given a budget constraint. Prices and incomes are uniformly sampled in the ranges $[1,10]$ and $[50,150]$, respectively, and we assume that the ground truth utility function is of Cobb-Douglas form, as given in equation \ref{eq:cobb-douglas}, with $k$ parameters. The optimal observations $\textbf{x}_i$ are computed by Sequential Least Squares Programming (SLSQP) \cite{kraft1988software} given the known values for prices $\textbf{p}_i$ and incomes $m_i$, for $i=1, \dots, N$. The dataset is then denoted by $D = \{\textbf{x}_i, \textbf{p}_i, m_i\}_{i=1}^N$. Generating the observations in this way ensures that the dataset is GARP-consistent as the agent is strictly utility maximising without deviations. 

\subsubsection{Utility Estimated by Cobb-Douglas} 

\begin{figure}[ht]
\begin{center}
\includegraphics[width=0.8\textwidth]{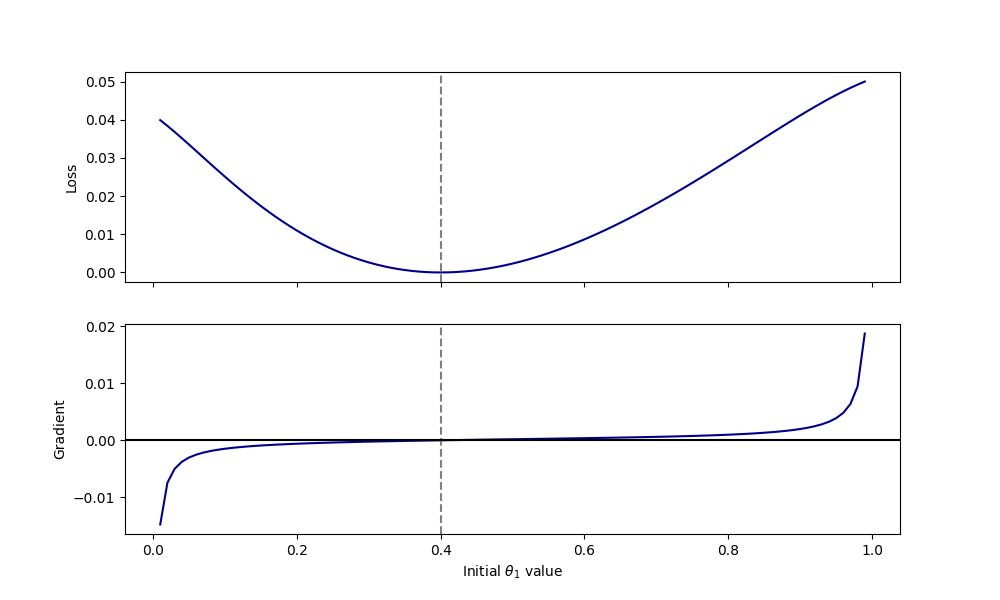}
\caption{Loss (top) and gradient (bottom) computed at varying initial values of $\hat{\theta}_1$, with true value at $\theta_1 = 0.4$ (dashed, grey). The loss is minimised and gradients are zero at the true value. Since the gradients cross the $x$-axis only once, the parameter is guaranteed to converge approximately at the true value with PEARL.}
\label{fig:cb-loss-grad}
\end{center}
\end{figure}

If the true utility which generated the observations is of CD form and a CD function is being fitted, it is shown that PEARL successfully recovers the true parameters. For this illustration, we generate a small sample size of $160$ observations, with bundles made up of $2$ goods ($k=2$) with CD parameters $\boldsymbol{\theta} = [0.4, 0.6]$. We chose $N=160$ because this corresponds to one batch of size $128$ after a $80-20$ train-test split, which removes any noise across batches. We have chosen these parameter values as an illustration; the same result applies across all valid parameter values. The loss and the gradients computed by the algorithm at various initial parameters $\hat{\boldsymbol{\theta}}$ are inspected. With the constraint that  $\sum_i^k \hat{\theta}_i = 1$, there is only $1$ parameter we need to estimate. We show that the loss and gradients differ with respect to different initial values for $\theta_1$. This is illustrated in Figure \ref{fig:cb-loss-grad}. Here, we see that for any initial value, the gradient update (by negative gradient) will ensure that the parameter values will converge at the true value, $\theta_1 =0.4$, which is where the loss is equal to $0$. 

\subsubsection{Utility Estimated by ICNN} \label{section:results-icnn}

\begin{figure}[ht] \label{fig:icnn-test-level-sets}
\begin{center}
\includegraphics[width=\textwidth]{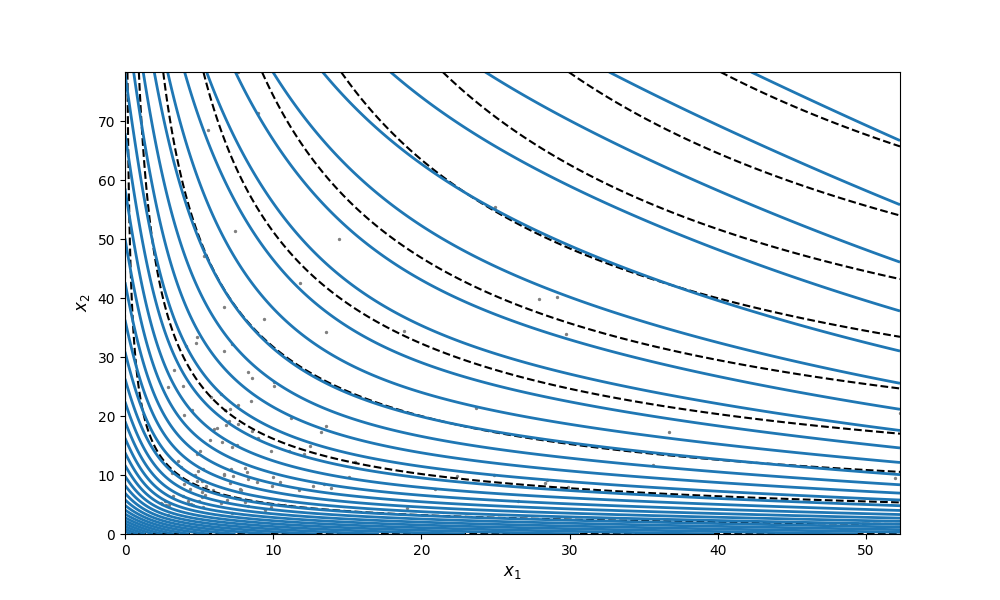}
\caption{Comparison of contours of fitted ICNN utility function (solid, blue lines) by PEARL and the ground truth CD utility function (black, dashed lines) with $\boldsymbol{\theta} = [0.4, 0.6]$. Observations shown in grey with $N=160$, $k=2$.}
\label{fig:icnn-level-sets}
\end{center}
\end{figure}

The ICNN, as described in section \ref{section:icnn}, is shown to be able to conform to the functional form of CD given a generated dataset. $N=160, k=2$ data points are generated with $\theta_1=0.4$ and an ICNN is fit with $3$ layers and $2$ hidden units at each layer. The fit of the ICNN utility function recovered via PEARL is illustrated by plotting the indifference curves (contours) of the utility function in Figure \ref{fig:icnn-level-sets}. The observations are indicated in grey, the ground truth contours in dashed black and the fitted model contours in blue. For a good fit, the fitted level sets should be parallel to the ground truth level sets since the utility function is ordinal and the shape of the level set is most important to obtain the optimal solution, given a budget constraint. 

\subsubsection{Computing Elasticities}

\begin{figure}
     \centering
     \begin{subfigure}[b]{0.495\textwidth}
         \begin{center}
             \includegraphics[width=\columnwidth]{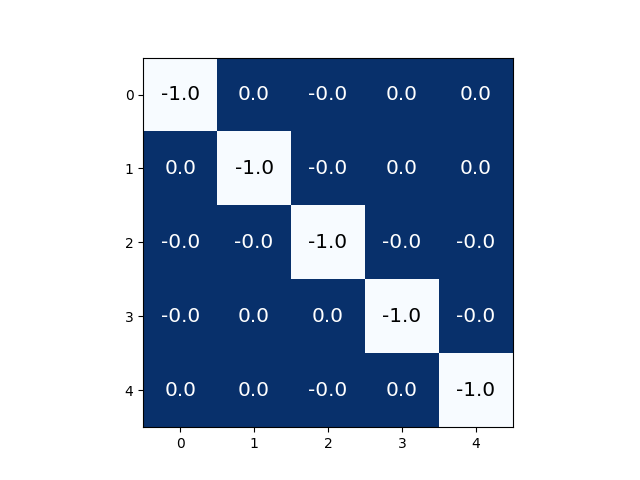}
         \caption{Price elasticities for $k=5$ from CD.}
         \label{fig:elas_5_cd}
         \end{center}
     \end{subfigure}
     \hfill
     \begin{subfigure}[b]{0.495\textwidth}
         \centering
         \includegraphics[width=\columnwidth]{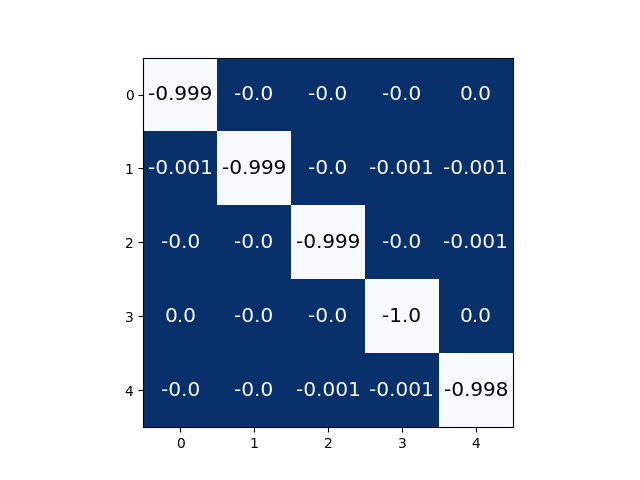}
         \caption{Price elasticities for $k=5$ from ICNN.}
         \label{fig:elas_5_icnn}
     \end{subfigure}
        \caption{Own and cross-price elasticities numerically estimated after CD (left) and ICNN (right) utility functions are trained by PEARL on $N=1600, k=5$. }
        \label{fig:elas-5}
\end{figure}

For the case where the Cobb-Douglas functional form is known and we get back the parameters, the elasticity computation is trivial, but the accuracy of the Algorithm \ref{algo:max-u} presented in section \ref{section:maximising-utility} can be tested. As such, the elasticities are computed by the method in section \ref{section:elasticities}. For $k$ products, the ground truth price elasticities are $e = -\textbf{I}$, where $\textbf{I}$ represents the $k \times k$ identity matrix. To illustrate, choose $k=5$ For the same data as described in section \ref{section:simulations}, with $k=5$, the elasticities estimated by the fitted utility function of CD and ICNN, $\hat{e}_{CD}$ and $\hat{e}_{ICNN}$, are presented in Figure \ref{fig:elas-5}.

\subsubsection{Estimating Demand from Utility}

Given the fitted utility functions using the PEARL algorithm, a counterfactual prediction of demand given a set of prices and income which form the budget constraint can be obtained by maximising the utility function, as in \ref{section:maximising-utility}. The RMSE is computed, given by
\begin{equation}
    RMSE(\hat{\textbf{x}}_i, \textbf{x}_i)= \sqrt{\frac{1}{N_s} \sum_i^{N_s} \sum_j^k (\hat{x}_i^j - x_i^j)^2}
\end{equation}
which refers to the difference in the actual predicted demand for each good in $\hat{\textbf{x}}_i$, where $\textbf{x}_i^j$ denotes the $j$-th element (good) of the observed bundle of goods for $i=1, \dots, N_s$ observations in the test set. A total of $N$ observations are sampled with an $80\%-20\%$ train-test split.  

We investigate the performance of each component of PEARL with an ablation study. We keep $k=2$ and $N=160$ and the network architecture is fixed as described in section \ref{section:icnn}. We compare RMSE scores for different activation functions in the ICNN fitted with PEARL, CD fitted with PEARL. We replace the concave activation functions with commonly used activation functions, ReLU \cite{agarap2019deeplearningusingrectified} and sigmoid \cite{hinton2012deep}, and remove the postivive constraints imposed on the weights. The activation functions we consider for the ICNN are the three concave activation functions, presented in section \ref{section:icnn}, \textit{concave-tanh}, \textit{concave-sigmoid} and \textit{concave-log}. The results are presented in Table \ref{table:pearl-ablation}.
\begin{table}[ht]
\caption{Demand prediction errors (RMSE) for simulated data generated with $k$ goods and $N$ observations using Cobb-Douglas. We compare a neural network regression with ReLU and sigmoid activation functions, PEARL fitted with non-input-concave neural networks and PEARL with ICNN and CD.}
\label{table:pearl-ablation}
\centering
\begin{center}
\resizebox{\textwidth}{!}{
\begin{tabular}{cc|cccccc}
\multirow{2}{*}{$k$} & \multirow{2}{*}{$N$}& PEARL (NN) & PEARL (NN)& PEARL (ICNN) & PEARL (ICNN) & PEARL (ICNN) & PEARL (CD)  \\
&  & \textit{ReLU} & \textit{sigmoid} & \textit{concave-tanh}& \textit{concave-sigmoid} & \textit{concave-log} & - \\ \midrule
$2$ & $160$ & 24.329 & 15.127 & 0.197 & 0.340 & 0.009 & \textbf{0.002}  \\ 
\bottomrule
\end{tabular}}
\end{center}
\end{table}

The performance is compared to that of machine learning regression methods for demand prediction, as shown in \cite{bajarietal2015}, who state that these machine learning methods outperform other forms of demand estimation. These include linear regression, LASSO, Random forest (RF), support vector machine (SVM), bagging, and $k$-nearest neighbours regression and XGBoost. The input is the normalised price (price divided by income) and the output is the amount consumed. The results are shown in Table \ref{table:models-rmse} where the root mean squared error (RMSE) for the train and test data is reported. We compare $k=2, N=160$ as before. We also now increase the number of goods with values to $5$ and $10$, with the number of observations set at $N=1600$. With a $80-20$ train-test split, we get $10$ batches of size $128$, which corresponds to approximately $3.5$ years of daily consumption data. 
\begin{table}[ht]
\caption{Demand prediction errors (RMSE) for simulated test data generated with $k$ goods and $N$ observations using Cobb-Douglas, fitted with PEARL ICNN and CD, compared to various machine learning methods (linear regression, LASSO, random forest (RF), support vector machines (SVM), Bagging, K-nearest neighbours (KNN), and XGBoost).}
\label{table:models-rmse}
\centering
\begin{center}
\resizebox{\textwidth}{!}{
\begin{tabular}{cc|ccccccccc}
\multirow{2}{*}{$k$} & \multirow{2}{*}{$N$} & \multirow{2}{*}{Linear} & \multirow{2}{*}{LASSO} & \multirow{2}{*}{RF} & \multirow{2}{*}{SVM} & \multirow{2}{*}{Bagging} & \multirow{2}{*}{KNN} & \multirow{2}{*}{XGBoost} & PEARL & PEARL \\
& & & & & & & & & (ICNN) & (CD) \\
\midrule
$2$ & $160$ & 14.118 & 19.613 &  4.647 & 14.855 & 6.821 & 10.655 & 1.984 & 0.009 & \textbf{0.002} \\ 
$5$ & $1600$ & 6.789 & 9.971 & 1.837 & 4.647 & 1.992 & 4.254 & 0.291 & 0.013 & \textbf{0.012} \\ 
$10$ & $1600$ & 4.984 & 7.383 & 2.764 & 3.899 & 2.945 & 4.310 & 0.299 & 0.052 & \textbf{0.027} \\ 
\bottomrule
\end{tabular}}
\end{center}
\end{table}

\begin{figure}
     \centering
     \begin{subfigure}[b]{0.495\textwidth}
         \centering
         \includegraphics[width=\columnwidth]{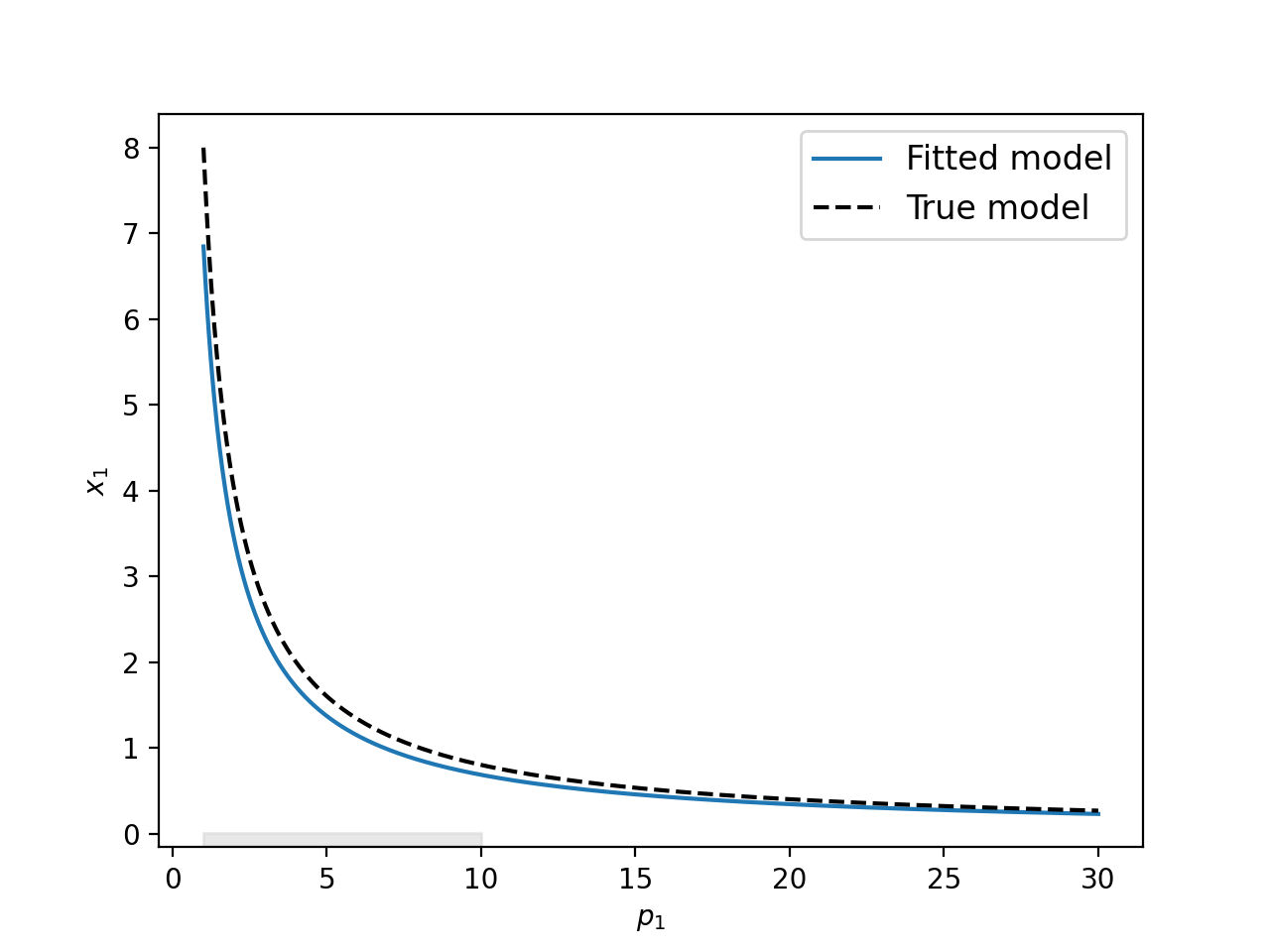}
         \caption{Demand for $x^1$.}
         \label{fig:demand_1}
     \end{subfigure}
     \hfill
     \begin{subfigure}[b]{0.495\textwidth}
         \centering
         \includegraphics[width=\columnwidth]{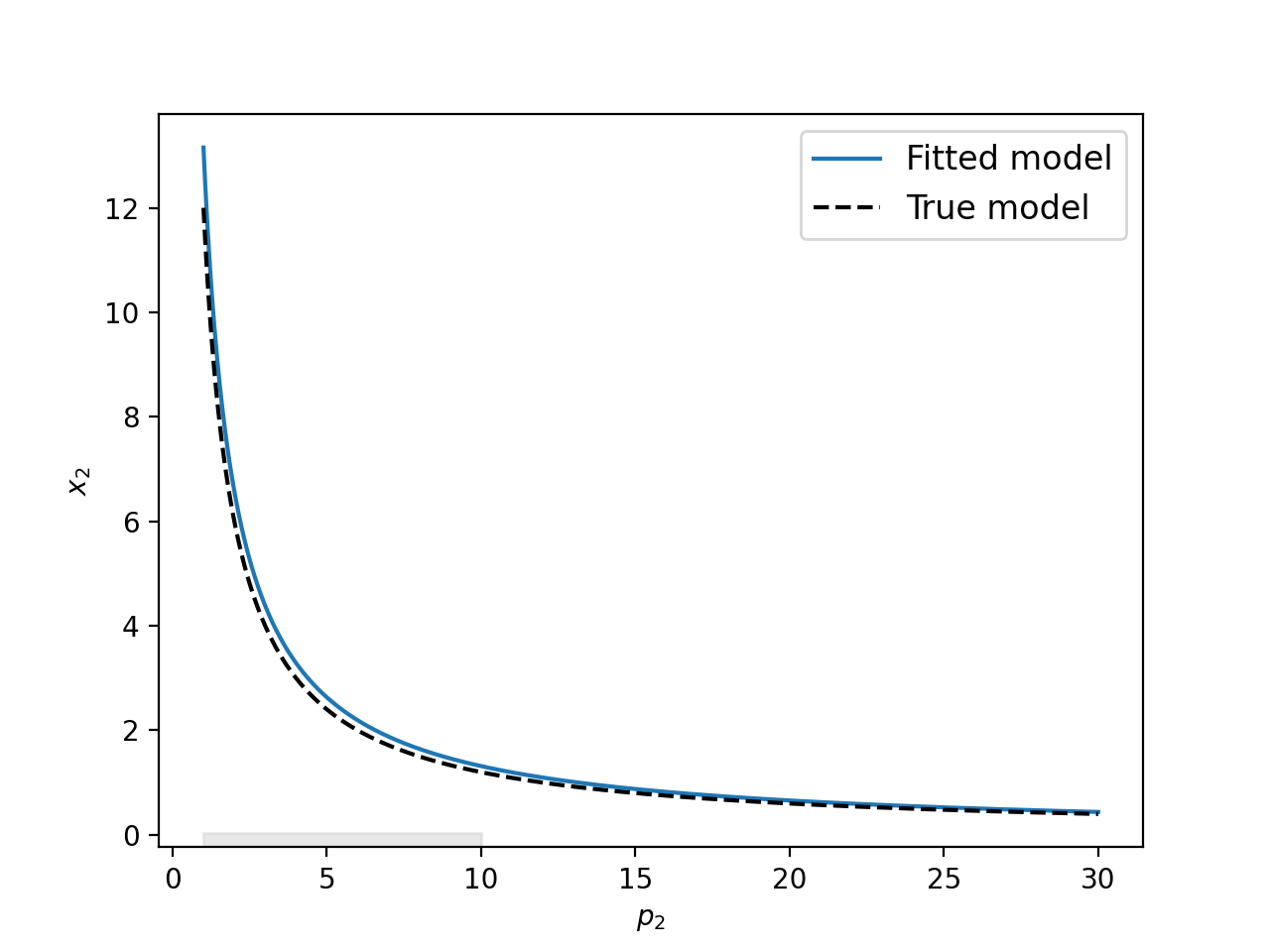}
         \caption{Demand for $x^2$.}
         \label{fig:demand_2}
     \end{subfigure}
        \caption{Illustration of comparison of demand functions for (a) $x^1$ and (b) $x^2$ between ground truth (black, dashed) and the demand obtained from maximising the fitted ICNN (blue). Prices for all the other goods have been fixed at $5$ and income at $20$. Observations are generated for prices between $1$ and $10$, as indicated by the grey bars.}
        \label{fig:icnn-demand}
\end{figure}

The demand function for one good is visualised, keeping the prices of the other goods fixed, which is compared to the analytical result from the ground truth. Notice that in the simulation, the price range is between $1$ and $10$; the plots illustrate that the function can extrapolate beyond the observed range. This is shown in Figure \ref{fig:icnn-demand}. 

\subsection{Simulations with Noise}

Next, perturbations are applied to the utility function or to the observations to simulate deviations from the utility-maximising consumer with constant preferences. Two cases are considered, inspired by \cite{Tipoe2021}. These are random utility, resulting from e.g. measurement error; and endogeneity, resulting in e.g. correlation between the parameters of the utility function and the prices observed. Using PEARL, the two specified utility functions of CD and ICNN form are fitted with the $\varepsilon$-adjustment, as described in section \ref{section:garp-consistency}, and compare them to machine learning regression methods. The results show that in each of the cases described, the PEARL algorithm outperforms the benchmark. 

\subsubsection{Random Utility}

Inconsistencies arising from measurement error or irrational behaviour leading to variation in the optimal consumption bundle denoted by $\textbf{x}$ can be simulated by introducing a random error $\eta$ as a perturbation. As such, the observed consumption choices are now 
\begin{equation}
    \hat{\textbf{x}}_i = \max_{\textbf{x}} \{U_{\theta}(\textbf{x}) \text{ s.t. } \textbf{p}_i^T\textbf{x} \leq m_i \} - |\eta|, 
\end{equation}
where $\eta \sim \mathcal{N}(\textbf{0}, \Sigma)$ with $\Sigma$ as the $k \times k$ positive semi-definite correlation matrix across the components of the consumption bundle. The observed income is then updated to $m_i = \textbf{p}_i^T\hat{\textbf{x}}_i$. This means that the observed consumption bundles may not be the ones that would have been obtained under the updated income value, thus resulting in deviations from the optimal and rational consumer. 

The RMSE values are reported for the counterfactual prediction of optimal consumption bundles resulting from the PEARL algorithm with $\varepsilon$-adjustment using CD and ICNN utility functions and their comparisons to machine learning regression method in Table \ref{table:rmse-random-noise}. We report the RMSE values for settings of $k=2, N=160$ and $k=5, N=1,600$. 

\begin{table}[ht]
\caption{Demand prediction errors (RMSE) for simulated test data generated with $k$ goods and $N$ observations using Cobb-Douglas with random noise, for fitted CD and ICNN utility functions using PEARL with $\varepsilon$-adjustment, compared to various machine learning methods (linear regression, LASSO, random forest (RF), support vector machines (SVM), Bagging, K-nearest neighbours (KNN), and XGBoost).}
\label{table:rmse-random-noise}
\centering
\begin{center}
\resizebox{\textwidth}{!}{
\begin{tabular}{cc|c|ccccccccc}
\multirow{2}{*}{k} & \multirow{2}{*}{N}& \multirow{2}{*}{$\varepsilon$} & \multirow{2}{*}{Linear} & \multirow{2}{*}{LASSO} & \multirow{2}{*}{RF} & \multirow{2}{*}{SVM} & \multirow{2}{*}{Bagging} & \multirow{2}{*}{KNN} & \multirow{2}{*}{XGBoost} & PEARL & PEARL \\
& & & & & & & & & & (ICNN) & (CD) \\
\midrule
$2$ & $160$ & 0.992 & 11.459 & 14.473 & 3.446 & 10.065 & 3.810 & 5.358 & 1.857 & \textbf{0.238} & 1.679 \\
$5$ & $1600$ & 0.844 & 7.370 & 9.926 & 2.571 & 5.435 & 2.829 & 5.018 & 2.107 & \textbf{0.830} & 2.664 \\ 
\bottomrule
\end{tabular}}
\end{center}
\end{table}

\subsubsection{Price Endogeneity}

In a system where prices are set dynamically as a response to (expected) demand, it is difficult to determine whether changes in consumption are due to changes in preferences or as a response to prices, or both. Such an environment is simulated by incorporating an error term which varies the parameters of the CD utility function, $\boldsymbol{\theta}$, and also varies the generated prices. As such, the prices are correlated with the parameters, which simulates the endogenous nature of prices. The system is described by the $k \times 1$ vector of prices $\textbf{p}$ being perturbed by a random error, $\eta$, such that the observed prices are now $\textbf{p}' = \textbf{p} + \eta$, where $\eta$ is a $k \times 1$ vector of random errors, with each element $j=1,\dots,k$ drawn from independently from $\eta^j \sim \mathcal{N}(0, 0.01)$ and with each element $j=1,\dots,k$ of the price vector drawn independently from $p^j \sim \mathcal{U}[1, 10]$. These errors are also applied to the $k \times 1$ vector of true parameters of the utility function, $\boldsymbol{\theta}$, giving $\boldsymbol{\theta}' = \boldsymbol{\theta} + \eta$. Applying the same perturbation to both the parameters and prices acts as a `demand shock' and leads to correlation between the price and the parameters of the utility function, thus simulating endogeneity. 
\begin{table}[ht]
\caption{Demand prediction errors (RMSE) for simulated test data with $k$ goods and $N$ observations, generated using Cobb-Douglas with endogeneity, for fitted CD and ICNN utility functions using PEARL with $\varepsilon$-adjustment, compared to various machine learning methods (linear regression, LASSO, random forest (RF), support vector machines (SVM), Bagging, K-nearest neighbours (KNN), and XGBoost).}
\label{table:rmse-endogeneity}
\centering
\begin{center}
\resizebox{\textwidth}{!}{
\begin{tabular}{cc|c|ccccccccc}
\multirow{2}{*}{$k$} & \multirow{2}{*}{$N$} & \multirow{2}{*}{$\varepsilon$} & \multirow{2}{*}{Linear} & \multirow{2}{*}{LASSO} & \multirow{2}{*}{RF} & \multirow{2}{*}{SVM} & \multirow{2}{*}{Bagging} & \multirow{2}{*}{KNN} & \multirow{2}{*}{XGBoost} & PEARL & PEARL \\
& & & & & & & & & & (ICNN) & (CD) \\
\midrule
$2$ & $160$ & 0.999 & 9.785 & 14.470 & 3.771 & 10.026 & 4.365 & 5.604 & 3.234 & 2.197 & \textbf{2.018} \\
$5$ & $1600$ & 0.909 & 7.223 & 10.267 & 3.987 & 5.628 & 4.193 & 5.537 & 3.846 & \textbf{3.065} & 4.530 \\ 
\bottomrule
\end{tabular}}
\end{center}
\end{table}
The RMSE values are reported for the counterfactual prediction of optimal consumption bundles resulting from the PEARL algorithm with $\varepsilon$-adjustment using CD and ICNN utility functions and their comparisons to machine learning regression method in Table \ref{table:rmse-endogeneity}. We report the RMSE values for settings of $k=2, N=160$ and $k=5, N=1,600$. 

\section{Conclusion}

Demand estimation is a challenging task due to the endogenous nature of price and demand. Existing methods which seek to identify the effect of price on demand have stringent IV requirements to determine exogenous variation and isolate the relationship between the two variables while accounting for the effect of demand shocks. These methods face difficulties in specifying IVs and the number required scales rapidly with the number of endogenous variables. Many commonly used methods which bypass this problem and do not rely on IVs are not able to identify the effect of price on demand because they do not address the problem of endogeneity. 

To address these challenges, we introduced PEARL, an algorithm that de-couples the effect of price on demand by explicitly modelling the consumer's decision-making process. PEARL is an algorithm that successfully recovers a representation of the underlying utility function of a utility-maximising consumer under a budget constraint, which generated observed consumption data as a result of the utility-maximising process. It uses economic theory from revealed preferences to obtain conditions for ensuring that the data set is consistent with a rational utility-maximising consumer. Under these conditions, a utility function that rationalises the observed data set is guaranteed to exist. In such a case, the algorithm obtains a representation of such a function by an iterative process. We also introduced a neural network architecture, the Input-Concave Neural Network (ICNN) that is concave with respect to the input with the use of one of the three concave activation functions: \textit{concave-tanh}, \textit{concave-sigmoid} and \textit{concave-log}. 

PEARL was tested on simulated data with and without noise, and real-world data. Given that the functional form is known, it has been shown that the true parameters which generated the data set can be recovered. We presented a neural network, the Input-Concave Neural Network (ICNN), which satisfies the conditions of a utility function through concave activation functions and constraints on the weights of the network. It can be used as a flexible model to recover the utility function. It has been shown to obtain small errors on very small sample sizes. These cases illustrated the performance on small sample sizes and a small number of goods. Further analysis is required to investigate its performance on larger sizes of bundles and samples, as well as more fine-tuned training and model optimisation per setting. 

A key limitation is that PEARL does not account for changing preferences or parameters of utility functions. The algorithm ensures that the data on which the model is trained are GARP consistent and does not utilise the calculated error, $\varepsilon$, aside from adjusting the observations. As such, the predictions resulting from the methodology will only ever be GARP-consistent (possibly with adjustment) with previous preferences as the change over time is not explicitly treated. 

Moreover, PEARL is built upon assumptions of well-behaved preferences, especially monotonicity and convexity, which means the consumer always wants to consume more goods and a mix of both goods is better. This is a strong assumption on consumer behaviour and while it may be reasonable to assume this in consumption settings such as in supermarkets, it may not be applicable to different scenarios, especially when decisions are more sparse over time. The result of these assumptions is the existence of a concave utility function under GARP-consistent observations. These assumptions may not hold in other scenarios and, as such, the utility function may not be concave. We have also assumed that the observations come from the same deterministic utility maximisation objective, as either the representative agent or homogenous agents; accounting for heterogeneity across individuals is an avenue for future work. 


\newpage
\bibliography{bib.bib}

\newpage
\appendix

\end{document}